# Applying Deep Learning to Detect Traffic Accidents in Real Time Using Spatiotemporal Sequential Data

Amir Bahador Parsa [a,1], Rishabh Singh Chauhan [a], Homa Taghipour [a], Sybil Derrible [a], Abolfazl (Kouros) Mohammadian [a]

[a] *University of Illinois at Chicago, 842 W Taylor Street, Chicago, IL 60607, USA*

**Abstract**

Accident detection is a vital part of traffic safety. Many road users suffer from traffic accidents, as well as their consequences such as delay, congestion, air pollution, and so on. In this study, we utilize two advanced deep learning techniques, Long Short-Term Memory (LSTM) and Gated Recurrent Units (GRUs), to detect traffic accidents in Chicago. These two techniques are selected because they are known to perform well with sequential data (i.e., time series). The full dataset consists of 241 accident and 6,038 non-accident cases selected from Chicago expressway, and it includes traffic spatiotemporal data, weather condition data, and congestion status data. Moreover, because the dataset is imbalanced (i.e., the dataset contains many more non-accident cases than accident cases), Synthetic Minority Over-sampling Technique (SMOTE) is employed. Overall, the two models perform significantly well, both with an Area Under Curve (AUC) of 0.85. Nonetheless, the GRU model is observed to perform slightly better than LSTM model with respect to detection rate. The performance of both models is similar in terms of false alarm rate.

[1] Corresponding author. Tel: +1 (312) 996-9840.
*Email addresses*: aparsa2@uic.edu (A. B. Parsa), rchauh6@uic.edu (R. S. Chauhan), htaghi2@uic.edu (H. Taghipour), derrible@uic.edu (S. Derrible), kouros@uic.edu (A. K. Mohammadian).

# 1. Introduction

Traffic accidents are a major concern in traffic safety, not only because they impact human lives (e.g., injuries and death), but also because they impose delay to road users, they increase fuel and energy consumption, they increase emission, and the list goes on. In fact, statistics shows that traffic accidents are responsible for approximately 60% of traffic congestion delays (*1*), thus directly increasing emissions and causing secondary damages (*2*). Moreover, traffic accidents are reported as a major human health issue with an estimated 1.25 million fatalities and 50 million injuries worldwide annually (*3*). Consequently, and despite an already abundant literature, the development of models to be able to rapidly and accurately detect accidents is paramount for traffic safety.

A wide range of techniques have been proposed and applied by researchers to detect the occurrence and estimate the duration of accidents. Generally, the literature on the topic can be divided into three general groups. The first group comprises studies that used simple methods and few data sources available since the 1970s, such as employing traffic flow theory and defining thresholds on traffic variables to detect accidents (*4*), or taking advantage of common distributions such as Poisson, negative binomial, Poisson-lognormal to analyze accidents (*5–7*). The second group comprises studies that used statistical and machine learning methods that are increasingly being used in transportation (*8–13*). Many researchers have employed Artificial Neural Network (ANN) (*14*), support vector machine (*15, 16*), probabilistic neural network (*17*), block clustering (*18*), random forest (*19*), pattern recognition (*20*), image processing techniques (*21*), and so on. Many of these techniques performed reasonably well while being partly able to cope with imbalanced or noisy data, especially when large volume of data is available. The last group of studies employs advanced techniques that are mostly used along with new sources of data and data fusion techniques. To this end, social media data (*22*), cellphone sensors data (*23*), vision-based sources of data (*24*), and traffic detector data (*25*) have been utilized, sometimes in models that used advanced techniques such as deep learning (*26*) and hybrid and extreme learning (*24, 27*). This group of studies generally achieves higher accuracy; however, they can be computationally more expensive than the other models, or they tend to require many more resources in terms of data storage and processing (*27*).

Thanks to their generally high performance and the availability of many data sources, deep learning techniques have been used in a large number of studies. More generally in the transport community, these techniques have also been exploited to accurately predict travel congestion evolution (*28*), predict peak-hour traffic condition (achieving 30% to 50% improvement) (*29*), improve prediction accuracy of traffic speed (*30, 31*), better predict traffic flow (*32*), detect precisely driver drowsiness based on latent facial features (*33*), and so on. In the field of traffic safety, specifically, researchers leverage deep learning techniques to predict traffic accidents (*34*), predict short-term crash risk (*35*), detect accidents (*26*), and predict injury severity of traffic accidents (*36*).

Moreover, a group of traffic safety studies has taken advantage of deep learning techniques to process vision-based data. In particular, Convolutional Neural Network (CNN) was found to work well with vision-based data (*37*). In an effort to prevent traffic accidents, several studies employed CNN models to detect drivers' distraction (*38*), drowsiness (*33*) and fatigue (*39*) based on facial features such as eye movement, blink rates, and so on. Shah et al. (*40*) utilized a specific type of CNN model called R-CNN to predict and analyze accidents using closed-circuit television traffic camera, and they were able to increase the object detection accuracy of their model by 8% and 6% by using Context Mining (CM) and Augmented Context Mining (ACM), respectively. In addition, Vu and Pham (*41*) used Empirical Deep CNN model to detect incidents on roads, which can work as a powerful tool for accident warning by exploiting video data. It is worth noting that this vision-based model can achieve detection accuracy of 97% for sunny weather condition, and 81% for night time or harsh weather conditions. Furthermore, Yuan et al. (*42*) used Convolutional Long Short-Term Memory (ConvLSTM) neural network model along with several sources of data including Satellite Images, Traffic Camera Data to predict accidents. This model performs considerably better than baseline models with respect to accuracy. Similarly, to the spatial analysis of traffic accidents, a method named spatiotemporal convolutional long short-term memory network (STCL-Net) was also proposed and outperformed several econometric and machine learning models such as autoregressive integrated moving average, random-parameter model, random-effects model, geographically weighted regression, CNN, LSTM, ANN, and gradient boosting regression tree (*35*).

As a deep learning technique, Recurrent Neural Networks (RNN) is known to work well with sequential data (i.e., time series) (*43*). Simple RNNs (*44*), however, are not as effective in learning long-term dependencies because of vanishing and exploding gradients (*43*). To remediate this problem, Long Short-Term Memory (LSTM) were developed with the aim to solve the problem of vanishing gradient (*43*, *44*), and they are also apt to model both short-term and long-term dependencies present in the data (*44*). The high prediction accuracy of LSTM model is thanks to its ability to memorize long-term historical data (*45*). As another type of RNN, Gated Recurrent Units (GRUs) (*43*) can also deal with dependencies at different time scales (*44*). In traffic safety, RNN techniques are particularly desirable since data is often in the form of time series (e.g., traffic conditions over a period of time) (*46*). In fact, RNN generally outperforms traditional techniques such as ANN and bayesian logistic regression using sequential data (*36*). Furthermore, Honglei et al. (*47*) achieved acceptable results to predict the risk of accidents by utilizing LSTM only on traffic time series data. They suggest integrating traffic data with other data sources such as human mobility, road characteristics, and special events to increase model accuracy. Finally, LSTM and GRU were found to perform best to monitor accidents based on structured data extracted from Twitter posts (*48*).

The objective of this study is to apply LSTM and GRU models to detect accidents using a set of real time data comprised of traffic time series, weather condition data, and congestion status data. Synthetic Minority Over-sampling Technique (SMOTE) is also employed to remediate problems linked with imbalanced data.

The remainder of the article is as follows. First, the next section describes the data used in this study and the feature extraction process used. Then, the methods and model validation procedure are explained. Subsequently, the results of the final models are presented and discussed. Finally, a conclusion is offered that summarizes the findings of the study and suggests potentially avenues for future research.

## 2. Data Description and Preprocessing

This study is conducted in the Chicago metropolitan area with a population of more than 9.5 million people. Detailed information on accidents' location and time are provided by the Illinois Department of Transportation (IDOT) for the period of December 2016 to December 2017. After cleaning the data, 241 accident cases are extracted for this study. In addition, 6,038 non-accident cases are defined randomly to represent normal traffic conditions at different times of day and for different links across the Chicago traffic network. **Figure 1** shows the location of the accidents in Chicago.

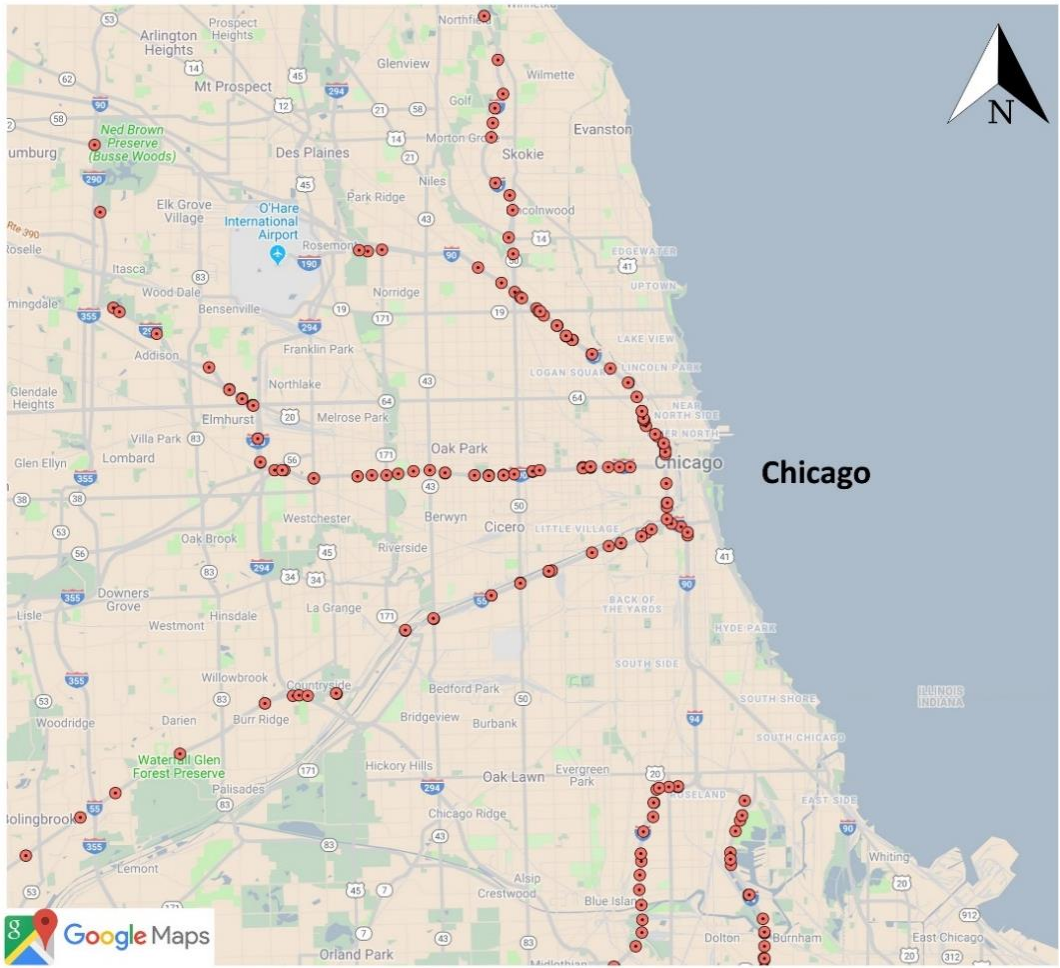

**Figure 1 Location of accidents in Chicago**

Any incident can significantly affect traffic conditions. **Figure 2** displays traffic conditions for accident and non-accident cases schematically. Essentially, each link possesses one loop detector upstream and one downstream of any location in a such way that, on average, the length of a link in the Chicago traffic network is about 0.6 mile (1 kilometer). For non-accident cases, it is expected that the traffic conditions do not vary considerably in successive time series between the upstream and downstream locations. In contrast, when an accident occurs, traffic conditions between the upstream and downstream loop detectors change rapidly because of the shockwaves caused by the accident, and this difference increases as time passes by.

The main source of data for this study integrates accident data with traffic temporospatial data. Traffic data includes traffic volume, occupancy, and speed, which are collected every 20 seconds by loop detectors. In this study, the data first are aggregated to 1 minute intervals from 5 minutes before to 5 minutes after an accident/non-accident time; the selection of five minutes stems from a different study by the authors of this article (*17*).

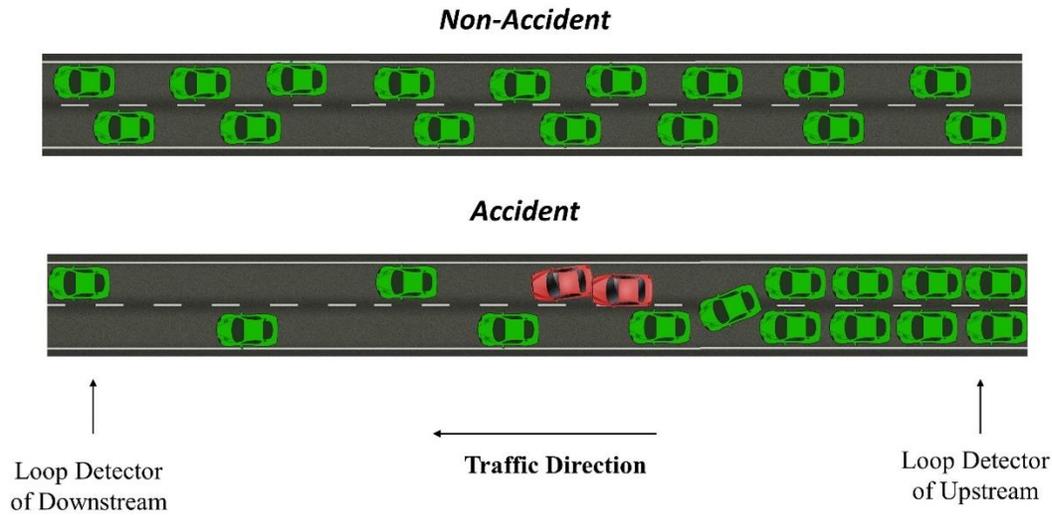

**Figure 2 Traffic condition in accident and non-accident cases**

**Figure 3** shows joint plots of traffic variables before and after accident/non-accident time for the upstream location. The green and red plots show non-accident and accident cases, respectively, and the darker areas on the graphs represent higher concentrations of data points. From **Figure 3**, for all three graphs of non-accident cases, the dense/darker areas occupy a much smaller space than those of accident cases. That is, in non-accident cases, traffic conditions are similar before and after non-accident time. In accident cases, however, traffic conditions differ significantly between before and after the time of accident, and the dense/darker areas occupy a larger space.

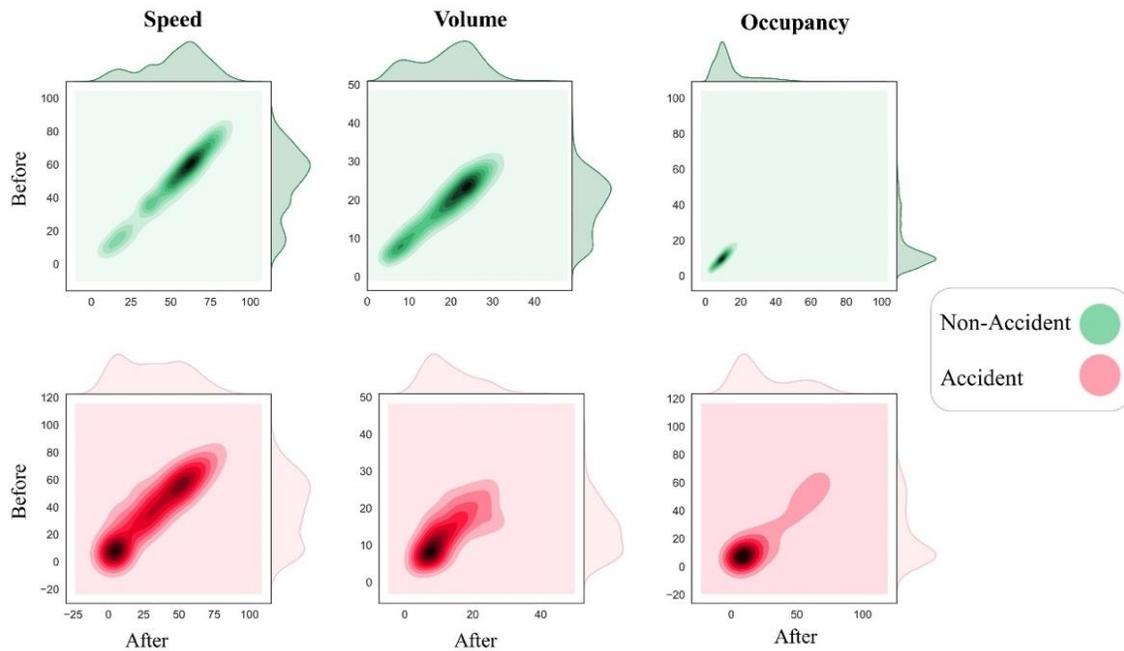

**Figure 3 Joint plot of traffic variables before and after accident/non-accident time, at upstream**

Along with traffic data, several other features are also generated to capture the impact of weather conditions and peak hours on the model. For the former, 94 different weather status collected at the Chicago's O'Hare and Midway international airports are aggregated into 4 status from 1 for sunny to 4 for stormy and harsh weather conditions. Regarding the latter, three dummy variables are generated for weekdays, morning peak hours, and evening peak hours. The final set of variables used in the model and their description are provided in **Table 1**.

**TABLE 1 Explanatory variables**

| Variables | Description | Mean |
|---|---|---|
| **Speed** | | |
| Downstream | Speed at downstream location, aggregated to 1 min intervals, from 5 minutes before to 5 minutes after an accident/non-accident case (mi/hr) | 50.43 |
| Upstream | Speed at upstream location, aggregated to 1 min intervals, from 5 minutes before to 5 minutes after an accident/non-accident case (mi/hr) | 50.83 |
| **Occupancy** | | |
| Downstream | Occupancy at downstream location, aggregated to 1 min intervals, from 5 minutes before to 5 minutes after an accident/non-accident case (%) | 14.94 |
| Upstream | Occupancy at upstream location, aggregated to 1 min intervals, from 5 minutes before to 5 minutes after an accident/non-accident case (%) | 15.29 |
| **Volume** | | |
| Downstream | Volume at downstream location, aggregated to 1 min intervals, from 5 minutes before to 5 minutes after an accident/non-accident case (veh/min) | 18.62 |
| Upstream | Volume at upstream location, aggregated to 1 min intervals, from 5 minutes before to 5 minutes after an accident/non-accident case (veh/mi) | 18.65 |
| **Weather Condition** | | |
| Weather | Ordinal variable from 1 for sunny to 4 for stormy weather conditions | 1.18 |
| **Day of Week** | | |
| Weekday | Dummy variable with 1 for weekday and 0 weekend | 0.71 |
| **Peak Hour** | | |
| Morning | Dummy variable with 1 for weekday morning peak hour and 0 otherwise | 0.12 |
| Evening | Dummy variable with 1 for weekday evening peak hour and 0 otherwise | 0.13 |

Number of observations: 6,279

## 3. Methodology

### 3.1. Model Development

Since the dataset is skewed towards non-accidents, SMOTE is used for balancing the data in the train set before it is fed into the models. In comparison with other oversampling techniques, SMOTE increases the number of data points from minority classes by generating new synthetic members instead of duplicating data points. For more information about this technique, the reader is referred to a study conducted by Chawla et al. (*49*). It is worth noting that the application of SMOTE on traffic imbalanced data has already been tested for this application (*17*).

Moreover, as is common with deep learning, all input variables are first scaled between 0 and 1 to avoid biases towards variables with large. Subsequently, the models output values in the range [0,1] that need to be converted to either 0 (i.e., no accident detected) or 1 (i.e., accident detected) by choosing a threshold value. In this study, the threshold value was selected by trying a range of values and by choosing the one that gives high accuracy and detection rate, and low false alarm rate in accident detection in the test dataset.

The LSTM and the GRU models were developed using Python (v3.7.3) through the Keras (v2.2.4) Deep Learning Library that uses TensorFlow (v2.0.0b0) in the backend. The details of these models are presented below.

*Long Short-Term Memory (LSTM)*
LSTM is a type of RNN (*43*) that involves cells and gates (*44*, *50*). A cell is an internal processing unit in LSTM (*44*). The gates are a combination of addition, multiplication, and nonlinearities, through which the information is added or removed from a cell (*50*). The LSTM consists of three gates, namely the forget gate, the input gate, and the output gate (*44*, *50*). LSTMs have a cell state, through which the main flow of information occurs (*50*). The role of the forget gate is to decide which information needs to be removed from the previous cell state h[t-1], while the input gate decides how much of the new state h[t] must be updated. Finally, the output gate decides the part of the state that has to be outputted (*44*). The behavior of these gates are controlled by a set of parameters that are trained through gradient descent (*44*). There are various approaches that can be applied to update the parameters (*44*). The current study utilizes the Adam approach.

The cell state is updated in the forward pass and the output is computed as shown in the difference equations below (*44*):

forget gate: $\quad \sigma_f[t] = \sigma(W_f \cdot x[t] + R_f \cdot y[t-1] + b_f)$ (1)

candidate state: $\quad \tilde{h}[t] = g_1(W_h \cdot x[t] + R_h \cdot y[t-1] + b_h)$ (2)

input gate: $\quad \sigma_u[t] = \sigma(W_u \cdot x[t] + R_u \cdot y[t-1] + b_u)$ (3)

cell state: $\quad h[t] = \sigma_u[t] \, (\cdot) \, \tilde{h}[t] + \sigma_f[t] \, (\cdot) \, h[t-1]$ (4)

output gate: $\quad \sigma_o[t] = \sigma(W_o \cdot x[t] + R_o \cdot y[t-1] + b_o)$ (5)

output: $\quad y[t] = \sigma_o[t] \, (\cdot) \, g_2(h[t])$ (6)

In these equations, x[t] is the input at time t, $\sigma(\cdot)$ is a sigmoid function, $g_1(\cdot)$ and $g_2(\cdot)$ denote the point wise nonlinear activation function, $(\cdot)$ denotes the entry wise multiplication between two vectors, $R_o$, $R_u$, $R_h$, and $R_f$ represents weight matrices of the recurrent connections, $W_o$, $W_u$, $W_h$, and $W_f$ are weight matrices for the inputs of LSTM cells, $b_o$, $b_u$, $b_f$, and $b_h$ are bias vectors (*44*).

The Adam optimizer was used in the models developed and the losses were computed as binary crossentropy. The hidden layers of the model consisted of sigmoid activation function. The hyperparameters of this model like the number of LSTM layers, the number of hidden layers, the number of neurons in each layer, the number of epochs, and the batch size were selected by manual hyperparametric tuning.

*Gated Recurrent Unit (GRU)*
GRU consists of only two gates, namely the update gate, and the reset gate (*43*). The function of the update gate is to decide how much of the current content needs to be updated, while that of the reset gate is to reset the memory, if it is closed (*44*).
The state equations of GRU are given as (*44*):

| | | |
|---|---|---|
| reset gate: | $r[t] = \sigma(W_r \cdot h[t-1] + R_r \cdot x[t] + b_r)$ | (7) |
| current state: | $h'[t] = h[t-1] \: (\cdot) \: r[t]$ | (8) |
| candidate state: | $z[t] = g(W_z \cdot h'[t-1] + R_z \cdot x[t] + b_z)$ | (9) |
| update gate: | $u[t] = \sigma(W_u \cdot h[t-1] + R_u \cdot x[t] + b_u)$ | (10) |
| new state: | $h[t] = (1-u[t]) \: (\cdot) \: h[t-1] + u[t] \: (\cdot) \: z[t]$ | (11) |

where $\sigma(\cdot)$ is a sigmoid function, $g(\cdot)$ represents point wise nonlinear activation function, $(\cdot)$ is the entry wise multiplication between two vectors, $W_u$, $W_z$, $W_r$, $R_u$, $R_z$, and $R_r$ are weight matrices, and $b_u$, $b_z$, and $b_r$ are bias vectors (*44*).

As in the case of LSTM model, the optimizer used for the GRU model was Adam, the losses were calculated as binary crossentropy, and the hidden layers in the model consisted of the sigmoid activation function. The hyperparameters of this model like the number of GRU layers, the number of hidden layers, the number of neurons in each layer, the number of epochs, and the batch size were chosen by manual hyperparametric tuning as well.

### 3.2. Model Evaluation

In order to evaluate performance of models, three measures are used in this study: accuracy, detection rate, and false alarm rate. The aim is to achieve high accuracy, high detestation rate, and low false alarm rate. These three measures are calculated through Equations 12 to 14.

$$\text{Accuracy} = \frac{Number\ of\ true\ reports}{Total\ number\ of\ cases} \times 100 \tag{12}$$

$$\text{Detection Rate} = \frac{Number\ of\ true\ accident\ reports}{Total\ number\ of\ accidents} \times 100 \tag{13}$$

$$\text{False Alarm Rate} = \frac{Number\ of\ false\ accident\ reports}{Total\ number\ of\ cases} \times 100 \tag{14}$$

## 4. Results

In this section, the results of the LSTM and GRU models based on the features displayed in the **Table 1** are presented. To train the model, the modeling dataset was split into a train set (65%) and a test set (35%) selected randomly from the dataset. **Table 2** provides the summary of model training procedure for both techniques.

TABLE 2 Model Summery

| Model Characteristics | Models | |
|---|---|---|
| | LSTM | GRU |
| Number of Neurons in the 1st layer | 30 | 50 |
| Number of Neurons in the 2nd layer | 50 | 70 |
| Number of Neurons in the 3rd layer | 60 | 90 |
| Number of Neurons in the 4th layer | 80 | 110 |
| Number of Neurons in the 5th layer | 50 | 100 |
| Number of Neurons in the 6th layer | 40 | 80 |
| Number of Neurons in the 7th layer | 30 | 60 |
| Number of Neurons in the 8th layer | 20 | 40 |
| Number of Neurons in the 9th layer | 10 | 20 |
| Number of Neurons in the output layer | 1 | 1 |
| Number of epochs | 2500 | 4000 |
| Number of Batches | 2000 | 2000 |
| Optimizer | Adam | Adam |
| Loss | Binary Crossentropy | Binary Crossentropy |
| TP | 62 | 63 |
| FP | 66 | 70 |
| FN | 22 | 21 |
| TN | 2048 | 2044 |

The GRU model achieved an accuracy, a detection rate, and a false alarm rate of 95.9%, 75%, and 3.2%, respectively. Comparatively, the LSTM model achieved as accuracy of 96%, a detection rate of 73.8%, and a false alarm rate of 3.0%. These results indicate that both LSTM and GRU models perform well to detect accidents, but the GRU performs slightly better. In order to better compare and present the performance of the two models, the Receiver Operating Characteristic (ROC) curve is plotted in **Figure 4**. From the figure, both LSTM and GRU output a similar Area Under Curve (AUC) of 0.85.

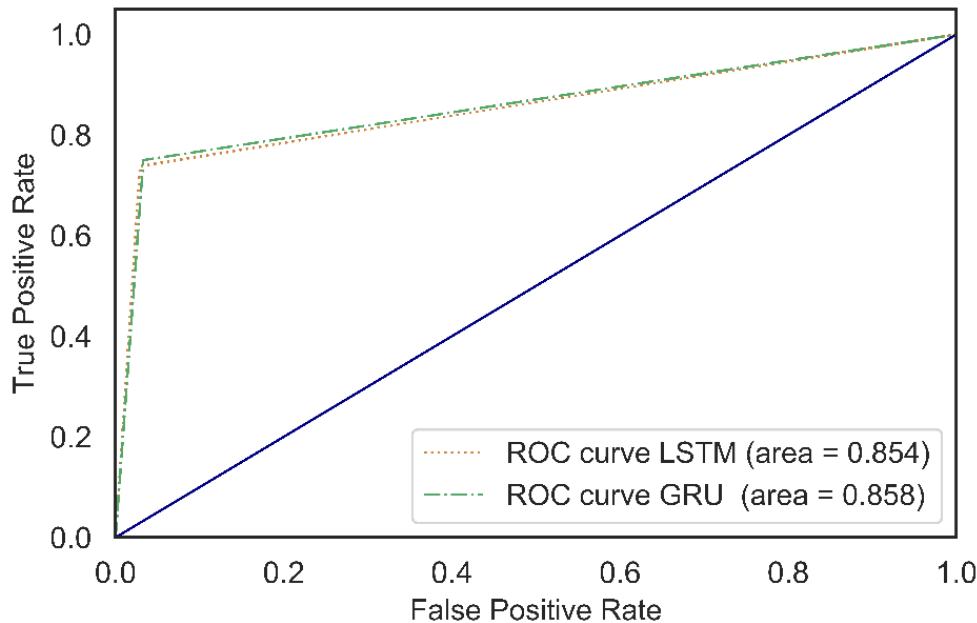

Figure 4 Receiver Operating Characteristic (ROC) curve of models

## 5. Summery and Conclusion

Traffic accidents is an important issue for traffic safety. Annually, 1.25 million deaths are caused by traffic accidents that are also costly in terms of money and time (i.e., traffic delay). Having access to real time data and advanced modeling techniques can help researchers and practitioners rapidly and accurately detect accidents. In this study, we considered 241 accident and 6,038 non-accident cases, and we utilized real time data along with LSTM and GRU (two deep learning techniques) to detect accidents. We also took advantage of SMOTE to balance the accident data. The results show that both models performed well, and that GRU performed marginally better.

LSTM and GRU are well-suited for this study since the data contains information about traffic characteristics at different time steps, and these techniques are able to deal with dependencies at different time scales. That is, GRU can capture both long-term and the short-term dependencies, and LSTM model is capable of memorizing long historical data. Accordingly, in accident detection modeling, a technique which can simultaneously achieve high detection rate and low false alarm rate is desirable. Based on the results, both GRU and LSTM models could achieve high detection rate with considerable low false alarm rates.

Given the availability of loop detector traffic data, detecting accident 5 min after they occur using real time loop detector data is satisfactory. That being said, the introduction of other datasets such as cellphone's GPS and accelerometer data could further improve the performance of the models so as to detect accidents even more rapidly.

**ACKNOWLEDGEMENTS**

This research was supported, in part, by the National Science Foundation (NSF) CAREER Award 1551731.

**REFERENCES**
1. Traffic Accident Management. *Federal Highway Administration*, 2014.
2. Traffic Incident Management. *Federal Highway Administration*, 2013.
3. Global Status Report on Road Safety 2015. *World Health Organization*, 2015.
4. H. J. Payne, and S. C. Tignor. Freeway Incident Detection Algorithms Based on Decision Tree with States. *Transportation Research Record*, 1978, pp. 30–37.
5. Jovanis, P. P., and H.-L. Chang. Modeling the Relationship of Accidents to Miles Traveled. *Transportation Research Record*, 1986.
6. Joshua, S. C., and N. J. Garber. Estimating Truck Accident Rate and Involvements Using Linear and Poisson Regression Models. *Transportation planning and Technology*, 1990.
7. Miaou, S.-P., and H. Lum. Modeling Vehicle Accidents and Highway Geometric Design Relationships. *Accident Analysis & Prevention*, 1993.
8. Arvin, R., M. Kamrani, and A. J. Khattak. How Instantaneous Driving Behavior Contributes to Crashes at Intersections: Extracting Useful Information from Connected Vehicle Message Data. *Accident Analysis & Prevention*, Vol. 127, 2019, pp. 118–133.
9. Azimi, G., A. Rahimi, H. Asgari, and X. Jin. Severity Analysis for Large Truck Rollover Crashes Using a Random Parameter Ordered Logit Model. *Accident Analysis & Prevention*, Vol. 135, 2020.
10. Arvin, R., M. Kamrani, and A. J. Khattak. The Role of Pre-Crash Driving Instability in Contributing to Crash Intensity Using Naturalistic Driving Data. *Accident Analysis & Prevention*, Vol. 132, 2019.
11. Movahedi, A., and S. Derrible. Interrelated Patterns of Electricity, Gas, and Water Consumption in Large-Scale Buildings. (under Review). *Sustainable Cities and Society*, 2020.
12. Parsa, A. B., K. Kamal, H. Taghipour, and A. (Kouros) Mohammadian. Does Security of Neighborhoods Affect Non-Mandatory Trips? A Copula-Based Joint Multinomial-Ordinal Model of Mode and Trip Distance Choices. *Transportation Research Board 98th Annual MeetingTransportation Research Board*, 2019.
13. Parsa, A. B., A. Movahedi, H. Taghipour, S. Derrible, and A. K. Mohammadian. Toward Safer Highways, Application of XGBoost and SHAP for Real-Time Accident Detection and Feature Analysis. *Accident Analysis & Prevention*, 2020.
14. Moggan Motamed. Developing A Real-Time Freeway Incident Detection Model Using Machine Learning Techniques. 2016.
15. Dong, N., H. Huang, and L. Zheng. Support Vector Machine in Crash Prediction at the Level of Traffic Analysis Zones: Assessing the Spatial Proximity Effects. *Accident Analysis and Prevention*, Vol. 82, 2015, pp. 192–198. https://doi.org/10.1016/j.aap.2015.05.018.
16. Mokhtarimousavi, S., J. C. Anderson, A. Azizinamini, and M. Hadi. Improved Support Vector Machine Models for Work Zone Crash Injury Severity Prediction and Analysis. *Transportation Research Record*, 2019.
17. Parsa, A. B., H. Taghipour, S. Derrible, and A. (Kouros) Mohammadian. Real-Time Accident Detection : Coping with Imbalanced Data. *Accident Analysis and Prevention*, Vol. 129, No. January, 2019, pp. 202–210. https://doi.org/10.1016/j.aap.2019.05.014.
18. Rahimi, A., G. Azimi, H. Asgari, and X. Jin. Clustering Approach toward Large Truck Crash Analysis. *Transportation Research Record*, 2019.
19. Ozbayoglu, A. M., G. Kucukayan, and E. Dogdu. A Real-Time Autonomous Highway Accident Detection Model Based on Big Data Processing and Computational Intelligence. 2017, pp. 1807–1813. https://doi.org/10.1109/BigData.2016.7840798.
20. Rong, Y. U., W. A. N. G. Guoxiang, J. Zheng, and W. A. N. G. Haiyan. Urban Road Traffic

Condition Pattern Recognition Based on Support Vector Machine. *ournal of Transportation Systems Engineering and Information Technology*, Vol. 13(1), 2013, pp. 130–136.
21. Marimuthu, R., A. Suresh, M. Alamelu, and S. Kanagaraj. DRIVER FATIGUE DETECTION USING IMAGE PROCESSING AND ACCIDENT PREVENTION. *International Journal of Pure and Applied Mathematics*, 2017.
22. Gu, Y., Z. Sean, and F. Chen. From Twitter to Detector : Real-Time Traffic Incident Detection Using Social Media Data. *Transportation Research Part C*, Vol. 67, 2016, pp. 321–342. https://doi.org/10.1016/j.trc.2016.02.011.
23. Fernandes, B., M. Alam, V. Gomes, J. Ferreira, and A. Oliveira. Automatic Accident Detection with Multi-Modal Alert System Implementation for ITS. *Vehicular Communications*, Vol. 3, 2016, pp. 1–11. https://doi.org/10.1016/j.vehcom.2015.11.001.
24. Vishnu, V. C. M., and M. R. R. Nedunchezhian. Intelligent Traffic Video Surveillance and Accident Detection System with Dynamic Traffic Signal Control. *Cluster Computing*, 2018, pp. 135–147. https://doi.org/10.1007/s10586-017-0974-5.
25. Chen, Q., X. Song, H. Yamada, and R. Shibasaki. Learning Deep Representation from Big and Heterogeneous Data for Traffic Accident Inference. *30th AAAI Conference on Artificial Intelligence*, 2016, pp. 338–344.
26. Zhang, Z., Q. He, J. Gao, and M. Ni. A Deep Learning Approach for Detecting Tra Ffi c Accidents from Social Media Data. *Transportation Research Part C*, Vol. 86, No. November 2017, 2018, pp. 580–596. https://doi.org/10.1016/j.trc.2017.11.027.
27. Chen, Y., Y. Yu, and T. Li. A Vision Based Traffic Accident Detection Method Using Extreme Learning Machine. *International Conference on Advanced Robotics and Mechatronics (ICARM)*, 2016, pp. 567–572. https://doi.org/10.1109/ICARM.2016.7606983.
28. Ma, X., H. Yu, Y. Wang, and Y. Wang. Large-Scale Transportation Network Congestion Evolution Prediction Using Deep Learning Theory. *PLOS ONE*, 2015, pp. 1–17. https://doi.org/10.1371/journal.pone.0119044.
29. Yu, R., Y. Li, C. Shahabi, U. Demiryurek, and Y. Liu. Deep Learning : A Generic Approach for Extreme Condition Traffic Forecasting. *SIAM International Conference on Data Mining*, 2017, pp. 777–785.
30. Wang, J., Q. Gu, J. Wu, G. Liu, and Z. Xiong. Traffic Speed Prediction and Congestion Source Exploration : A Deep Learning Method. *2016 IEEE 16th International Conference on Data Mining (ICDM)*, 2016, pp. 499–508. https://doi.org/10.1109/ICDM.2016.0061.
31. Prediction, N. S. Learning Traffic as Images : A Deep Convolutional Neural Network for Large-Scale Transportation Network Speed Prediction. *Sensors*, 2017. https://doi.org/10.3390/s17040818.
32. Koesdwiady, A., R. Soua, and F. Karray. Improving Traffic Flow Prediction With Weather Information in Connected Cars: A Deep Learning Approach. *IEEE Transactions on Vehicular Technology*, Vol. 65, No. 12, 2016, pp. 9508–9517. https://doi.org/10.1109/TVT.2016.2585575.
33. Dwivedi, K., K. Biswaranjan, and A. Sethi. Drowsy Driver Detection Using Representation Learning. *2014 IEEE International Advance Computing Conference (IACC)*, 2014, pp. 995–999. https://doi.org/10.1109/IAdCC.2014.6779459.
34. Yuan, Z., X. Zhou, and T. Yang. Hetero-ConvLSTM : A Deep Learning Approach to Traffic Accident Prediction on Heterogeneous Spatio-Temporal Data. *Applied Data Science Track Paper*, 2018, pp. 984–992.
35. Bao, J., P. Liu, and S. V Ukkusuri. A Spatiotemporal Deep Learning Approach for Citywide Short-Term Crash Risk Prediction with Multi-Source Data. *Accident Analysis and Prevention*, Vol. 122, No. October 2018, 2019, pp. 239–254.

https://doi.org/10.1016/j.aap.2018.10.015.
36. Sameen, M. I., and B. Pradhan. Applied Sciences Severity Prediction of Traffic Accidents with Recurrent Neural Networks. *Applied Sciences*, 2017. https://doi.org/10.3390/app7060476.
37. Wang, Y., D. Zhang, Y. Liu, B. Dai, and L. Hay. Enhancing Transportation Systems via Deep Learning : A Survey. *Transportation Research Part C*, Vol. 99, No. October 2018, 2019, pp. 144–163. https://doi.org/10.1016/j.trc.2018.12.004.
38. Kim, W., H. K. Choi, B. T. Jang, and J. Lim. Driver Distraction Detection Using Single Convolutional Neural Network. *International Conference on Information and Communication Technology Convergence*, 2017.
39. Mandal, B., L. Li, G. S. Wang, and J. Lin. Towards Detection of Bus Driver Fatigue Based on Robust Visual Analysis of Eye State. *IEEE Transactions on Intelligent Transportation Systems*, 2016, pp. 545–557.
40. Shah, A. P., J. B. Lamare, T. Nguyen-Anh, and A. Hauptmann. CADP: A Novel Dataset for CCTV Traffic Camera Based Accident Analysis. *15th IEEE International Conference on Advanced Video and Signal Based Surveillance*, 2018, pp. 1–9.
41. Vu, N., and C. Pham. Traffic Incident Recognition Using Empirical Deep Convolutional Neural Networks Model. *Context-Aware Systems and Applications, and Nature of Computation and Communication*, 2017, pp. 90–99.
42. Liu, J., J. Wan, D. Jia, B. Zeng, D. Li, C. Hsu, and H. Chen. Version : Accepted Version Article : High-Efficiency Urban-Traffic Management in Context-Aware Computing and 5G Communication. *IEEE Communications Magazine*, 2017.
43. Pal, A., and P. K. S. Prakash. Practical Time Series Analysis: Master Time Series Data Processing, Visualization, and Modeling Using Python. 2017.
44. Bianchi, F. M., E. Maiorino, M. C. Kampffmeyer, A. Rizzi, and R. Jenssen. Recurrent Neural Networks for Short-Term Load Forecasting: An Overview and Comparative Analysis. 2017.
45. Tian, Y., and L. Pan. Predicting Short-Term Traffic Flow by Long Short-Term Memory Recurrent Neural Network. *2015 IEEE International Conference on Smart City/SocialCom/SustainCom (SmartCity)*, 2015, pp. 153–158. https://doi.org/10.1109/SmartCity.2015.63.
46. Wang, X., and M. Abdel-Aty. Temporal and Spatial Analyses of Rear-End Crashes at Signalized Intersections. *Accident Analysis & Prevention*, 2006, pp. 1137–1150.
47. Ren, H., Y. Song, J. Wang, Y. Hu, and J. Lei. A Deep Learning Approach to the Citywide Traffic Accident Risk Prediction. *2018 21st International Conference on Intelligent Transportation Systems*, 2018, pp. 3346–3351.
48. Fatkulin, T., N. Butakov, A. Krikunov, and D. Voloshin. Accident Monitoring Framework Based on Online Social Network Sensing. *Procedia computer science*, 2017, pp. 278–287.
49. Chawla, N., K. W. Bowyer, L. O. Hall, and W. P. Kegelmeyer. SMOTE: Synthetic Minority over-Sampling Technique. *Journal of Artificial Intelligence Research*, Vol. 16, 2002, pp. 321–357. https://doi.org/10.1613/jair.953.
50. Skansi, S. Introduction to Deep Learning: From Logical Calculus to Artificial Intelligence. 2018.